\documentclass[10pt,letterpaper]{article}
\usepackage[top=0.85in,left=2.75in,footskip=0.75in]{geometry}

\usepackage[T1]{fontenc}
\usepackage{algorithm}
\usepackage{algorithmic}
\usepackage{amsthm,amsmath,amssymb,eucal}
\usepackage{graphicx, color}
\usepackage{booktabs}
\usepackage{pgffor}
\usepackage[misc]{ifsym}

\usepackage[shortlabels]{enumitem}
\usepackage{float}
\usepackage{xcolor}
\usepackage[aboveskip=1pt,labelfont=bf,labelsep=period,justification=raggedright,singlelinecheck=off]{caption}

\usepackage{indentfirst}
\usepackage[hidelinks]{hyperref}

\usepackage{makecell}
\usepackage{comment}
\usepackage{svg}
\usepackage{multirow}

\begin{document}
\vspace*{0.2in}
\begin{flushleft}
{\Large
\textbf{Predicting O-GlcNAcylation sites in mammalian proteins with transformers and RNNs trained with a new loss function}
}
\newline
\\
Pedro Seber\textsuperscript{1*},
\\
\bigskip
\textbf{1} Department of Chemical Engineering, Massachusetts Institute of Technology, Cambridge, MA 02139, USA
\\
\bigskip
* pseber@mit.edu
\end{flushleft}
\section*{Abstract}
O-GlcNAcylation, a subtype of glycosylation, has the potential to be an important target for therapeutics, but methods to reliably predict O-GlcNAcylation sites had not been available until 2023; a 2021 review correctly noted that published models were insufficient and failed to generalize. Moreover, many are no longer usable. In 2023, a considerably better recurrent neural network (RNN) model was published. This article creates improved models by using a new loss function, which we call the weighted focal differentiable MCC. RNN models trained with this new loss display superior performance to models trained using the weighted cross-entropy loss; this new function can also be used to fine-tune trained models. An RNN trained with this loss achieves state-of-the-art performance in O-GlcNAcylation site prediction with an F\(_1\) score of 38.88\% and an MCC of 38.20\% on an independent test set from the largest dataset available.

\section*{Author Summary}
Glycosylation is the attachment of a sugar or sugars to a protein. This natural process has many essential functions in organisms. O-GlcNAcylation is a subtype of glycosylation that has been identified as having high potential for therapeutics. However, no method for systematic identification of O-GlcNAcylation sites exists. Models have been used for this task, but a 2021 review correctly noted that many models have low performances when predicting on data outside their training sets, making them unsuitable for practical use. A 2023 work made a considerable improvement in models for O-GlcNAcylation, leading to the first usable model for O-GlcNAcylation site prediction. We create improved models, which have a greater ability to identify O-GlcNAcylation sites, in turn better assisting the development of related therapeuticals and increasing our understanding of the biology behind this process. 

We also create a new loss function for classification models. Machine-learning models are optimized by minimizing a loss function, which represents the difference between the model's current performance and a perfect performance. The most widely used loss function only indirectly optmizes a model's performance metrics, but our loss function can directly optmize important metrics, leading to better performance, including for models made for other (non-glycosylation) tasks.

\section*{Introduction} \label{introduction_section}
Glycosylation, a co- and post-translational modification, occurs when glycans are added to proteins. O-linked glycosylation occurs when a glycan is added to an oxygen of an amino acid (usually serine or threonine in mammalian proteins). O-GlcNAcylation, a subtype of O-linked glycosylation, occurs when the first glycan added is an N-Acetylglucosamine (abbreviated GlcNAc) \cite{Schjoldager-etal-2020}. Unlike other forms of glycosylation, O-GlcNAcylation may be viewed similarly to phosphorylation, for no other glycans are further added to a site after GlcNAc and the process is catalyzed solely by two enzymes, OGT and OGA. O-GlcNAcylation is important functionally and structurally \cite{Schjoldager-etal-2020,Chang-etal-2020}; conversely, incorrect O-GlcNAcylation or its improper removal is associated with multiple diseases such as cancers \cite{Shi-etal-2022}, infections \cite{Chang-etal-2020}, and heart failure \cite{Umapathi-etal-2021}. According to recent research, O-GlcNAcylation can be a powerful target for therapeutics \cite{Zhu-and-Hart-2021}, further emphasizing its relevance.

While O-GlcNAcylation is essential for human health and can be critical for biotherapeutics, challenges still exist. Because O-GlcNAcylation is complete after the addition of a single glycan, the problem of predicting glycan distributions in a glycosylation site does not exist; thus, the main predictive task is to determine where and when (if ever) an amino acid will be O-GlcNAcylated. As summarized by Ref.\ \cite{Seber-and-Braatz-2023_2}, this task is challenging for ``multiple reasons'', which include ``a low frequency of events (only about 2\% of S/T [serine/threonine] sites are O-GlcNAcylated) and a lack of a motif to guide predictive efforts''. Moreover, Ref.\ \cite{Seber-and-Braatz-2023_2} also note that ``the effects of neighboring amino acids likely influence whether an S/T is O-GlcNAcylated'', a fact corroborated by their interpretability studies done with their final model, but many previous works use model architectures that cannot directly take into account any type of sequential information.

The pioneer classifier machine-learning model for this challenging task is YinOYang (YoY) \cite{Gupta-and-Brunak-2002}, which also had one of the highest performances until the work of Ref.\ \cite{Seber-and-Braatz-2023_2}, first published in 2023. However, until at least 2021, O-GlcNAcylation site prediction models had insufficient performance to help advance research in the area \cite{Mauri-etal-2021}. Specifically, Ref.\ \cite{Mauri-etal-2021} found that no published model until then could achieve a precision \(\ge\) 9\% on a medium-sized independent dataset, suggesting that models to predict the location of O-GlcNAcylation sites failed to generalize successfully in spite of the high metrics achieved with their training data. The models evaluated by Ref.\ \cite{Mauri-etal-2021} also have low F\(_1\) scores and Matthews Correlation Coefficients (MCCs), another sign that their performance is insufficient. Ref.\ \cite{Seber-and-Braatz-2023_2} used recurrent neural network (RNN) models (specifically, long short-term memory [LSTM] models) to predict the presence of O-GlcNAc sites from mammalian protein sequence data. That work used two different datasets (from Refs.\ \cite{Mauri-etal-2021,Wulff-Fuentes-etal-2021}), the latter of which is among the largest O-GlcNAcylation datasets available. These RNNs obtained vastly superior performance to the previously published models, achieving an F\(_1\) score more than 3.5-fold higher and an MCC more than 4.5-fold higher than the previous state of the art. Furthermore, Shapley values were used to interpret the predictions of that RNN model through the sum of simple linear coefficients \cite{Shapley-1951}. These predictions with Shapley values maintained most of the performance of the original RNN models (Supplemental data of Ref.\ \cite{Seber-and-Braatz-2023_2}) and provided interpretability for the first time in a publication on data-driven O-GlcNAcylation site prediction models.

Conversely, two other models that were also released recently, LM-OGlcNAc-Site \cite{Pokharel-etal-2023} and O-GlcNAcPRED-DL \cite{Hu-etal-2024}, failed to obtain high metrics in independent test sets despite their complex architectures (an ensemble of transformers pre-trained for proteins in LM-OGlcNAc-Site's case, and a CNN+LSTM model in O-GlcNAcPRED-DL's case) and the direct use of some structural information. These two models still cannot reach precision \(\ge\) 9\% on independent test sets, and an analysis by Ref.\ \cite{Seber-and-Braatz-2023_2} (Table 4 of that work) noted that their model had a 2-fold performance improvement over O-GlcNAcPRED-DL, despite the fact that the latter was operating with a few advantages.

This work creates improved O-GlcNAcylation prediction models by using three methods: transformer encoder models directly trained on O-GlcNAcylation data, finetuning of a pre-trained protein transformer \cite{Brandes-etal-2022}, both of which had lower performances than LSTM models, and transformers and RNNs trained with a new loss function, which we call the weighted focal differentiable MCC. This novel loss function aims to directly improve the models' MCCs, attaining superior results relative to those obtained with the weighted cross-entropy loss function. LSTMs trained with this new loss function reach state-of-the-art O-GlcNAcylation site prediction results, and the improvement in MCC compared to the cross-entropy loss function is statistically significant. 5-fold cross-validation is employed in the model training, ensuring the reliability of reported metrics and predictions. The improved model and novel loss function generated by this study are provided as open-source software, allowing the reproducibility of the work, the retraining of the models as additional or higher quality O-GlcNAcylation data become available, the use of the models to further improve the understanding and applications of O-GlcNAcylation, and the use of this novel loss function to improve models in other relevant tasks, as already done in Ref.\ \cite{Seber-and-Braatz-2025}, for example.

\section*{Results}
\subsection*{Transformers Have Good Predictive Power, but Are Inferior to RNNs} \label{results_ProtBERT}
\begin{figure}[h!]
    \hspace{-12ex}
    \centerline{
        \includegraphics[width=1.85\textwidth]{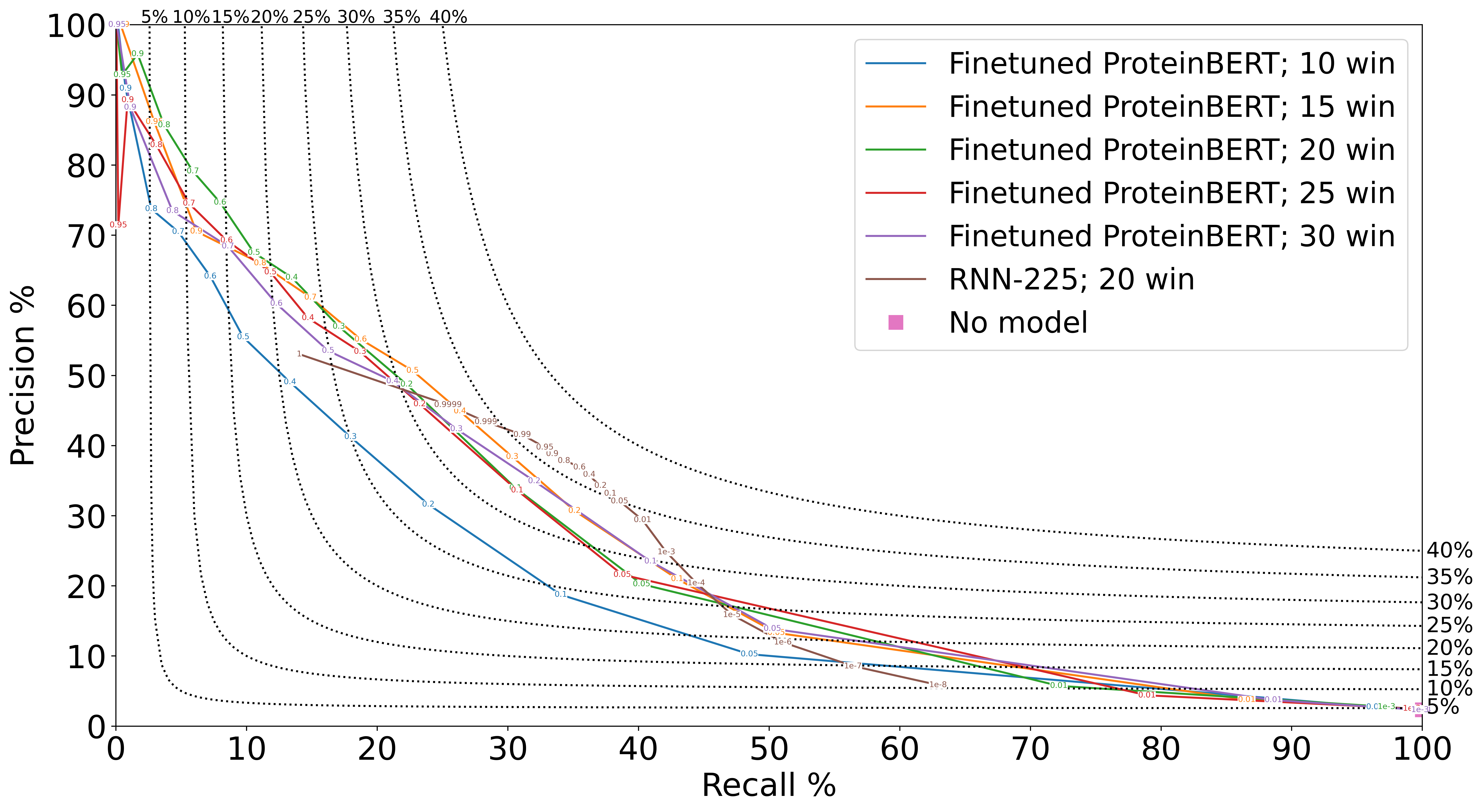} }
    \caption{P-R curves for the first seed of O-GlcNAcylation site prediction finetuned ProteinBERT models tested on the modified dataset of Ref.\ \cite{Wulff-Fuentes-etal-2021}. The black dotted lines are isolines of the F$_1$ score, as labeled on the top and right sides of the plot. The number after the semicolon represents the window size on each side of the central S/T. The numbers on the curves represent the minimum threshold for a site to be considered positive. The ``RNN-225'' model \cite{Seber-and-Braatz-2023_2} is the previous state-of-the-art model and is included for comparison. ``No Model'' is the baseline (threshold = 0).}
    \label{PR-curve_finetune}
\end{figure}

First, transformer encoder models were trained directly on this task. Although these models surpassed multiple models previously reported in the literature, they were inferior to the RNN of Ref.\ \cite{Seber-and-Braatz-2023_2} (Section \ref{Appendix_transformers}). Next, the ProteinBERT transformer was finetuned on this task. These finetuned transformers performed better than the transformer encoders directly trained on the O-GlcNAcylation data, but slightly worse than the RNN of Ref.\ \cite{Seber-and-Braatz-2023_2} (Table \ref{Metrics_finetune} and Fig.\ \ref{PR-curve_finetune}). Nevertheless, these finetuned transformers achieve good performance, surpassing multiple other models in the literature, including other transformer-based models (such as the previously mentioned LM-OGlcNAc-Site \cite{Pokharel-etal-2023}, which has a 2.3-fold lower MCC on an independent test set). The performance of finetuned ProteinBERT models would be state-of-the-art if not for the RNN of Ref.\ \cite{Seber-and-Braatz-2023_2} and the RNNs trained in this work.

Surprisingly, the finetuned ProteinBERT models' performances began to saturate and be maximized with a window size of 15, whereas other model types had this saturation and maximization only with a window size of 20. As with the other models, a window size of 25 or 30 did not improve performance in a statistically significant way relative to a window size of 20. It is possible that transformers are an ill-suited architecture for this task, potentially due to the amount of data available, which is why we train primarily RNNs in the following sections. The work of Ref.\ \cite{Brandes-etal-2022}, which trained ProteinBERT, also noted that LSTMs surpassed transformer models in multiple non-glycosylation-related tasks of biological relevance, corroborating this suitability hypothesis.

\begin{table*}[h!]
    \caption{Performance metrics (in \%) for the highest F$_1$ score point of finetuned ProteinBERT models tested on the modified dataset of Ref.\ \cite{Wulff-Fuentes-etal-2021}. WS is the window size on each side of the central S/T. The ``RNN-225'' model \cite{Seber-and-Braatz-2023_2} is the previous state-of-the-art model and is included for comparison. Baseline (threshold = 0) metrics are precision = 2.44\%, F\(_1\) score = 4.76\%, and MCC = 0\%.}
    \label{Metrics_finetune}
    \centerline{\begin{tabular}{| c | c c c c c | c |}
    \hline
    \multirow{2}{*}{Metric} & \multicolumn{5}{c |}{Finetuned ProteinBERT (This Work)} & \thead{RNN-225 \\ (Ref.\ \cite{Seber-and-Braatz-2023_2})} \\
    & WS = 10 & WS = 15 & WS = 20 & WS = 25 & WS = 30 & WS = 20 \\
    \hline
    Recall (\%)      & 23.09\(\pm\)2.51 & 30.45\(\pm\)3.91 & 29.23\(\pm\)2.44 & 32.20\(\pm\)1.97 & 31.48\(\pm\)0.80 & 35.47 \\
    Precision (\%)   & 31.75\(\pm\)5.31 & 36.78\(\pm\)6.61 & 36.25\(\pm\)2.72 & 35.98\(\pm\)4.27 & 36.67\(\pm\)4.64 & 36.90 \\
    F$_1$ Score (\%) & 26.41\(\pm\)0.74 & 32.79\(\pm\)1.07 & 32.24\(\pm\)0.68 & 33.86\(\pm\)1.74 & 33.76\(\pm\)1.46 & 36.17 \\
    MCC (\%)         & 25.42\(\pm\)1.18 & 31.72\(\pm\)1.43 & 31.05\(\pm\)0.45 & 32.46\(\pm\)1.98 & 32.48\(\pm\)1.85 & 34.57 \\
    \hline
    \end{tabular} }
\end{table*}

\subsection*{RNNs + Our Loss Function Have State-of-the-Art Predictive Performance} \label{results_diff-MCC}
\begin{figure*}[h]
    \hspace{-12ex}
    \centerline{
        \includegraphics[width=1.85\textwidth]{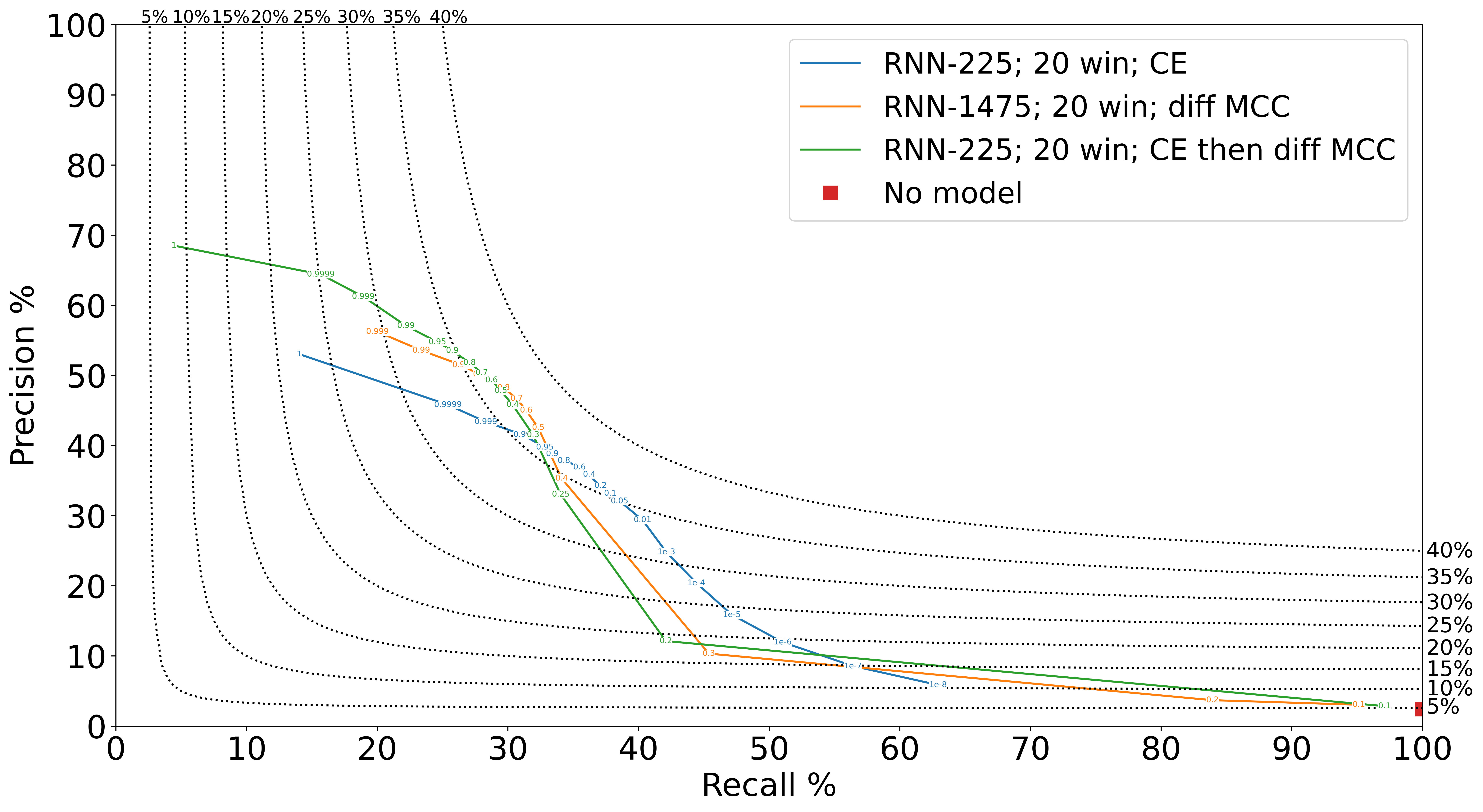} }
    \caption{P-R curves for O-GlcNAcylation site prediction single-cell RNN models tested on the modified dataset of Ref.\ \cite{Wulff-Fuentes-etal-2021}. Labels, curve descriptions, and baseline metrics are as in the captions of Table \ref{Metrics_finetune} and Fig.\ \ref{PR-curve_finetune}.}
    \label{PR-curve_single-layer}
\end{figure*}

With the weighted focal differentiable MCC loss, we then trained an RNN model and also finetuned the RNN model from Ref.\ \cite{Seber-and-Braatz-2023_2}. Ref.\ \cite{Seber-and-Braatz-2023_2} trained only single-cell RNNs, and so this constraint has been maintained for the first set of RNNs trained in this work. The RNN trained directly with our novel loss function (``RNN-1425; diff MCC'') achieves an F\(_1\) score = 37.03\% and an MCC = 36.58\% (Table \ref{Metrics_single-layer} and Fig.\ \ref{PR-curve_single-layer}). The relative and absolute increases in MCC are higher than those for the F\(_1\) score, highlighting the effectiveness of the weighted focal differentiable MCC loss, as it directly optimizes the MCC. To show that the weighted focal differentiable MCC loss can also be used for finetuning, the RNN model from Ref.\ \cite{Seber-and-Braatz-2023_2}, which was first trained using weighted cross-entropy, was finetuned using the weighted focal differentiable MCC loss (``RNN-225; CE \(\rightarrow\) diff MCC''). This model initially followed the optimal hyperparameters (including model size) found through cross-validation using the weighted cross-entropy loss, but the learning rate and loss hyperparameters are changed to the optimal values found through cross-validation with the weighted focal differentiable MCC loss function when that loss function is used for fine-tuning. This fine-tuning also leads to a model that is better than the previous state-of-the-art from Ref.\ \cite{Seber-and-Braatz-2023_2} (Fig.\ \ref{PR-curve_single-layer} and Table \ref{Metrics_single-layer}), reaching an F\(_1\) score = 36.52\% and an MCC = 36.01\%, values slightly lower than that of the model trained only with the weighted focal differentiable MCC loss.

\begin{table*}[h!]
    \caption{Performance metrics (in \%) for the highest F$_1$ score point of single-cell RNN models tested on the modified dataset of Ref.\ \cite{Wulff-Fuentes-etal-2021}. Labels and baseline metrics are as in the caption of Table \ref{Metrics_finetune}.}
    \label{Metrics_single-layer}
    \centerline{\begin{tabular}{| c | c | c | c |}
    \hline
    Metric & \thead{RNN-225; CE \\ (Ref.\ \cite{Seber-and-Braatz-2023_2})} & \thead{RNN-1425; diff MCC \\ (This Work)} & \thead{RNN-225; CE \(\rightarrow\) diff MCC \\ (This Work)} \\
    \hline
    Best Threshold   &  0.60 &  0.70 &  0.40 \\
    Recall (\%)      & 35.47 & 30.67 & 30.35 \\
    Precision (\%)   & 36.90 & 46.70 & 45.84 \\
    F$_1$ Score (\%) & 36.17 & 37.03 & 36.52 \\
    MCC (\%)         & 34.57 & 36.58 & 36.01 \\
    \hline
    \end{tabular} }
\end{table*}

There is a significant difference between the optimal hyperparameters selected through the weighted cross-entropy and weighted focal differentiable MCC loss functions. Compared to the optimal values found through the weighted cross-entropy loss, the weighted focal differentiable MCC loss has an optimum at higher model hidden sizes (15 vs. 240 for transformers; 225 vs. 1425 for RNNs), higher batch sizes (256 vs. 512 for transformers; 32 vs. 128 for RNNs), lower learning rates (1\(\times 10^{-3}\) vs. 1\(\times 10^{-4}\)), and lower class weights (15 vs. 3 for transformers; 30 vs. 3 for RNNs).

\begin{figure*}[h]
    \hspace{-12ex}
    \centerline{
        \includegraphics[width=1.85\textwidth]{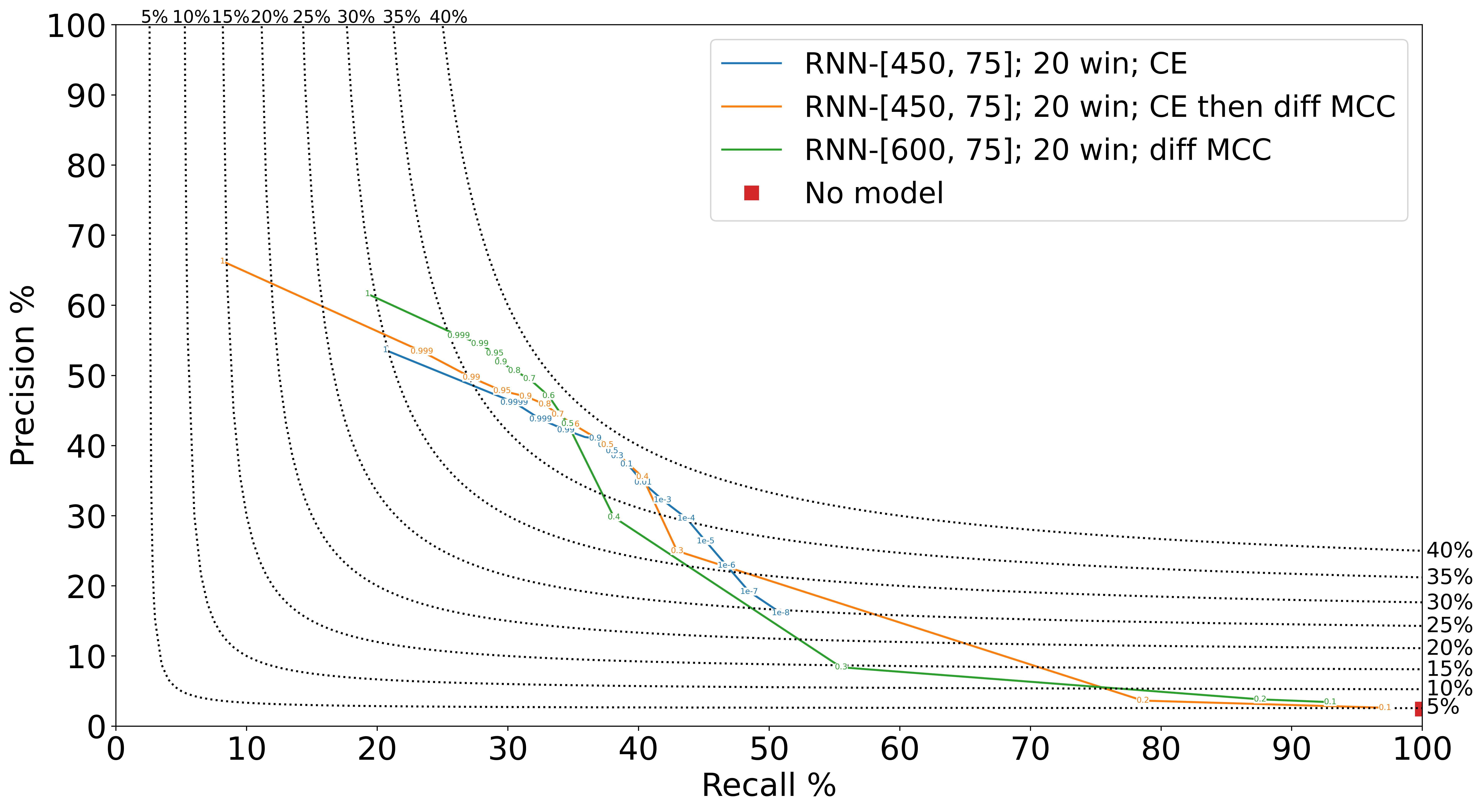} }
    \caption{P-R curves for O-GlcNAcylation site prediction two-cell RNN models tested on the modified dataset of Ref.\ \cite{Wulff-Fuentes-etal-2021}. Labels, curve descriptions, and baseline metrics are as in the captions of Table \ref{Metrics_finetune} and Fig.\ \ref{PR-curve_finetune}.}
    \label{PR-curve_multi-layer}
\end{figure*}

RNNs with multiple stacked cells may perform better than single-cell RNNs. To verify this hypothesis, RNN models with 2, 3, and 4 LSTM cells are tested (as per Section ``\nameref{MnM-models}''). As with the single-cell RNNs, three models are created: one trained only with the weighted CE loss, one trained only with the weighted focal differentiable MCC loss, and one first trained with the weighted CE loss, then fine-tuned with the weighted focal differentiable MCC loss. These models all surpass their single-cell equivalents (Table \ref{Metrics_multi-layer} and Fig.\ \ref{PR-curve_multi-layer}). The model trained with the weighted CE loss (``RNN-[450,75]; CE'') displays the greatest relative improvement over its single-cell version, reaching an F\(_1\) score = 38.73 \% and an MCC = 37.31\%. Like with the single-cell RNNs, fine-tuning with the weighted focal differentiable MCC loss after training with the weighted CE leads to improvements in the model performance; this fine-tuned model (``RNN-[450,75]; CE \(\rightarrow\) diff MCC'') reaches an F\(_1\) score = 38.83\% and an MCC = 37.33\%. While the performance improvement of fine-tuning is smaller in this scenario, these improved results again confirm the efficacy of fine-tuning with the weighted focal differentiable MCC loss. Finally, a model trained directly with the weighted focal differentiable MCC loss (``RNN-[600,75]; diff MCC'') achieves the best results, reaching an F\(_1\) score = 38.88\% and an MCC = 38.20\%. The increase in MCC is considerably higher, further corroborating the benefits of this new loss function discussed above. Overall, the ``RNN-[600,75]; diff MCC'' model displays a 7.5\% increase in F\(_1\) score and a 10.5\% increase in MCC over the ``RNN-255; CE'' model of Ref.\ \cite{Seber-and-Braatz-2023_2}.

\begin{table*}[h!]
    \caption{Performance metrics (in \%) for the highest F$_1$ score point of two-cell RNN models tested on the modified dataset of Ref.\ \cite{Wulff-Fuentes-etal-2021}. Labels and baseline metrics are as in the caption of Table \ref{Metrics_finetune}.}
    \label{Metrics_multi-layer}
    \centerline{\begin{tabular}{| c | c | c | c |}
    \hline
    Metric & \thead{RNN-[450,75]; CE \\ (This Work)} & \thead{RNN-[450,75]; CE \(\rightarrow\) diff MCC \\ (This Work)} & \thead{RNN-[600,75]; diff MCC \\ (This Work)} \\
    \hline
    Best Threshold   &  0.90 &  0.50 &  0.60 \\
    Recall (\%)      & 36.69 & 37.62 & 33.11 \\
    Precision (\%)   & 41.02 & 40.11 & 47.10 \\
    F$_1$ Score (\%) & 38.73 & 38.83 & 38.88 \\
    MCC (\%)         & 37.31 & 37.33 & 38.20 \\
    \hline
    \end{tabular} }
\end{table*}

The models with the highest average cross-validation F\(_1\) score all had only two layers and second LSTM layers with 75 neurons. As with the single-cell models, the optimal hyperparameters for models trained with the weighted focal differentiable MCC loss included higher model hidden sizes (600 vs.\ 450 for CE), lower learning rates (5\(\times 10^{-4}\) vs.\ 1\(\times 10^{-3}\)), and lower class weights (2 vs.\ 20). Surprisingly, both models had the same optimal batch size (128), and the hidden size and class weights for the model trained with the weighted CE loss moved towards those from the model trained with the weighted focal differentiable MCC loss.

Finally, we verified that our splits for the cross-validation and test sets do not bias the models by performing five-fold nested cross-validation, following the procedure in Section ``\nameref{MnM-dataset}'', on the ``RNN-[600,75]; diff MCC'' model. The models generated by each fold had very similar performance to each other (Fig.\ \ref{PR-curve_nested_validation} and Table \ref{Metrics_nested_validation_MCC}), highlighting the robustness of that architecture and loss function to changes in the data. In addition, a comparison of the nested validation results obtained with the weighted focal differentiable MCC loss function (Table \ref{Metrics_nested_validation_MCC}) and those obtained with the CE loss function (Table \ref{Metrics_nested_validation_CE}) show that the increases in MCC brought by the use of the diffMCC loss function are statistically significant (p = 0.0115) over multiple test sets, corroborating the success of our novel loss function in improving models by directly optimizing their MCCs.

\section*{Discussion} \label{discussion_section}
This work generates transformer encoder (Section \ref{Appendix_transformers}), finetuned transformer (Section ``\nameref{results_ProtBERT}''), and RNN (Section ``\nameref{results_diff-MCC}'') models to predict O-GlcNAcylation sites based on protein sequences using a modified version of an extensive public dataset \cite{Wulff-Fuentes-etal-2021}. These models are also compared to the state-of-the-art RNNs trained by Ref.\ \cite{Seber-and-Braatz-2023_2}. As the weighted cross-entropy loss function only indirectly optimizes the classification performance metrics, we hypothesized that a function that could directly improve the F\(_1\) score and MCC can lead to better-performing models. We develop the weighted focal differentiable MCC loss to address this question (Section ``\nameref{MnM-loss}''; Algorithm \ref{diffMCC}). This function directly optimizes the MCC, yielding better classification models.

This new loss function was used to generate a state-of-the-art model for O-GlcNAcylation site prediction, reaching an F\(_1\) score = 38.88\% and an MCC = 38.20\% (Fig.\ \ref{PR-curve_multi-layer} and Table \ref{Metrics_multi-layer}). Overall, this is a 7.5\% increase in F\(_1\) score and a 10.5\% increase in MCC over the ``RNN-255; CE'' model of Ref.\ \cite{Seber-and-Braatz-2023_2}. This improved model can assist in the discovery of novel O-GlcNAcylation sites, in turn expanding our knowledge of glycan biology and enabling the discovery of O-GlcNAcylation-related drugs \cite{Zhu-and-Hart-2021}. As the identification of O-GlcNAcylation sites is a laborious, specialized, and time-consuming process, the use of models and computational tools can accelerate research in the field.

Furthermore, the creation of a new loss function for classification tasks (Algorithm \ref{diffMCC}) can enable the training of better models, as classification is ubiquitous in many scientific, engineering, and social fields. This loss function improved models in a statistically significant way (p = 0.0115). These improvements occur not just for models used for O-GlcNAcylation-related tasks, as evidenced by the successful use of our weighted focal differentiable MCC loss function in other works \cite{Seber-and-Braatz-2025}.

The code used in this work is freely and publicly available, allowing the reproduction, improvement, and reuse of this work. It is simple to install and run the best RNN model to predict O-GlcNAcylation sites based on the local protein sequence. Instructions are provided in Sections \ref{Appendix_reproduction} and \ref{Appendix_using_model} of the supplemental data or the README at \href{https://github.com/PedroSeber/O-GlcNAcylation_Prediction}{https://github.com/PedroSeber/\\O-GlcNAcylation\_Prediction}.

\section*{Methods}
\subsection*{Dataset} \label{MnM-dataset}
The data used to train the model was first generated by Ref.\ \cite{Wulff-Fuentes-etal-2021} and was modified by Ref.\ \cite{Seber-and-Braatz-2023_2}. The modified version is available at \\
\href{https://github.com/PedroSeber/O-GlcNAcylation_Prediction/blob/master/OVSlab_allSpecies_O-GlcNAcome_PS.csv}{https://github.com/PedroSeber/O-GlcNAcylation\_Prediction/blob/master/\\OVSlab\_allSpecies\_O-GlcNAcome\_PS.csv}. This dataset contains 558,168 unique serine/threonine (S/T) sites from mammalian proteins. Out of these, 13,637 (2.44\%) are O-GlcNAcylated. The same procedure as in Ref.\ \cite{Seber-and-Braatz-2023_2} was employed to filter homologous and isoform proteins, a sequence selection based on a window size of 5 amino acids (AAs) on each side of the central S/T (11 AAs total), even for the larger windows. 20\% of each dataset is separated for testing, with the remaining 80\% used for cross-validation with five folds; a procedure also previously done in Ref.\ \cite{Seber-and-Braatz-2023_2}. To ensure that the choice of train/cross-validation and test sets is not biased and to highlight the robustness of our architecture and novel loss function, this work also performs nested validation with the best model, similarly to Ref.\ \cite{Seber-and-Braatz-2023_2}. In each nested validation round, 20\% of the data are used as a test set and the remaining is used for hyperparameter selection via 5-fold cross-validation (as above).

\subsection*{Transformer and RNN Models} \label{MnM-models}
Transformer encoder and RNN (specifically, LSTM) models are constructed using PyTorch \cite{PyTorch} and other Python packages \cite{Numpy,scikit-learn,pandas}. Moreover, the ProteinBERT transformer model \cite{Brandes-etal-2022} was also finetuned for this task. Details on the hyperparameters used for each model can be found in Section \ref{appendix_hyperparameters}. The best combination of hyperparameters is determined by a grid search, and the combination with the highest cross-validation average F\(_1\) score is selected. Finally, the performance on an independent test set of a model using that combination of hyperparameters is reported. Multiple thresholds are used to construct a Precision-Recall (P-R) curve, and the F\(_1\) scores and MCCs of the models are analyzed, allowing for a throughout evaluation of the best model's performance. F\(_1\) scores and MCCs are used for evaluation because they are widely regarded as the best performance metrics for classification models, as discussed in Section 2.3 of Ref.\ \cite{Seber-and-Braatz-2023_2} and references within.

\subsection*{Loss Functions} \label{MnM-loss}
The cross-entropy (CE) loss is the most common loss function for classification tasks, and its weighted variant is suitable for unbalanced data. Despite its widespread use and efficacy, the CE loss has some important limitations. First, the CE loss is distinct from the evaluation metrics used. Users of classification models may desire to maximize the F\(_1\) score or MCC metrics, but the model with the lowest CE loss is not necessarily the one with the highest metrics. Moreover, a model can lower its CE loss by manipulating the predictions of samples that are already correctly classified. For an extreme example, consider a model that predicts 0.8 for all positive samples and 0.2 for all negative samples and a second model that predicts 0.99 for all positive samples and 0.01 for all negative samples. Both models perform perfect classification, yet the second model will have a lower CE loss. In a less extreme scenario, a model may improve its CE loss by simply manipulating the probabilities of easy-to-classify samples, but such manipulations would not improve real metrics such as the F\(_1\) score or MCC.

There are multiple ways to remedy these issues. Ref.\ \cite{Lin-etal-2017} created the Focal Loss, which provides a correction factor \(\gamma\) that reduces the effect of increasing the certainty of a correct prediction, forcing the model to improve on difficult-to-classify samples to lower the focal loss. Ref.\ \cite{Berman-etal-2017} created the Lov{\'a}sz loss, which is an optimizable form of the Jaccardi Index and is claimed to perform better than the CE loss due to its better categorizing of small objects and lower false negatives. Other losses not evaluated in this work include the Baikal loss of Ref.\ \cite{Gonzalez-and-Miikkulainen-2020}, the soft cross entropy of Ref.\ \cite{Ilievski-and-Feng-2017}, the sigmoid F\(_1\) of Ref.\ \cite{Benedict-etal-2022}, the work of Ref.\ \cite{Lee-etal-2021}, and the MCC loss of Ref.\ \cite{Abhishek-and-Hamarneh-2021}. In particular, the approaches of this work resemble those of Ref.\ \cite{Lee-etal-2021} and Ref.\ \cite{Abhishek-and-Hamarneh-2021}, but these references did not use weighting or focal transforms, and the definitions for the differentiable F\(_1\) and MCC loss functions used in these works are slightly different from the definitions in this work.

The standard formulations of evaluation metrics (such as the F\(_1\) score or MCC) are not differentiable, so they cannot be directly used as loss functions. However, it is possible to modify them for differentiability. The method chosen to add differentiability is treating the true / false categories as prediction probabilities instead of binary values. For example, if the model predicts \(p_t = 0.7\) for a positive (negative) sample, that sample would increase the true positives (negatives) by 0.7 and the false negatives (positives) by 0.3. This modification can be considered as a differentiable MCC loss function. The focal modification of Ref.\ \cite{Lin-etal-2017} can be added to this loss function by modifying \(p_t\) to \((p_t)^\gamma\), which leads to the focal differentiable MCC loss function. Finally, it is possible to add class weighting: by multiplying the true positives and false negatives by a scalar W, it is possible to prioritize predictions on the positive class (W > 1) or negative class (W < 1). All the above changes combined lead to the \textbf{weighted focal differentiable MCC} loss function (Algorithm \ref{diffMCC}). We choose to optimize the MCC in particular because of its various desirable properties, especially in the context of imbalanced data \cite{Chicco-2017,Chicco-and-Jurman-2023}.

\begin{algorithm}[h]
   \caption{(binary) weighted focal differentiable MCC}
   \label{diffMCC}
\begin{algorithmic}
   \STATE {\bfseries Input:} predictions \(pred\) and targets \(y\); hyperparameters \(W\) (weight) and \(\gamma\) (focal loss factor) \\
   \STATE Define the following variables: \\
   \# The asterisk (*) represents multiplication. \textit{.sum()} is as per the NumPy/PyTorch function. \\
   \# \(p_t\) is the probability of a correct prediction, independently of whether the true target is positive or negative.
   \STATE \(p_t = pred*y + (1 - pred)*(1-y)\)
   \STATE TP = \( ((p_t)^\gamma * y).sum() * W\)
   \STATE FN = \( ((1-p_t)^\gamma * y).sum() * W\) \\
   \# \(W\) is used only if y = 1, that is, if the sample is in the positive class.
   \STATE FP = \( ((1-p_t)^\gamma * (1-y)).sum() \)
   \STATE TN = \( ((p_t)^\gamma * (1-y)).sum() \)
   \STATE Calculate the MCC as normal
   \STATE \textbf{return -MCC} \# Negative for maximization
\end{algorithmic}
\end{algorithm}

\section*{Declaration of competing interest}
The authors declare that they have no known competing financial interests or personal relationships that could have appeared to influence the work reported in this paper.

\section*{Data availability statement}
The raw data, code used to process these data, and models underlying this article are available in a GitHub repository at \href{https://github.com/PedroSeber/O-GlcNAcylation_Prediction}{https://github.com/\\PedroSeber/O-GlcNAcylation\_Prediction}.

\bibliographystyle{plos2025}
\bibliography{References}
\clearpage
\setcounter{page}{1}
\renewcommand{\thepage}{S\arabic{page}}
\makeatletter

\setcounter{section}{0}
\renewcommand{\thesection}{S\arabic{section}} 
\renewcommand{\thesubsection}{S\arabic{section}.\arabic{subsection}} 
\makeatletter

\section{Reproducing the Models and Plots} \label{Appendix_reproduction}
The transformer models can be recreated by downloading the datasets and running the \texttt{transformer\_train.py} file with the appropriate flags (run \texttt{python transformer\_train.py -{}-help} for details).

The RNN models can be recreated by downloading the datasets and running the \texttt{ANN\_train.py} file (modified from the original of Ref.\ \cite{Seber-and-Braatz-2023_2}) with the appropriate flags (run \texttt{python ANN\_train.py -{}-help} for details).

The plots can be recreated by running the \texttt{make\_plot2.py} file (modified from the original of Ref.\ \cite{Seber-and-Braatz-2023_2}) with the appropriate flag as an input (\texttt{python make\_plot2.py finetune} for Fig.\ \ref{PR-curve_finetune}, \texttt{python make\_plot2.py single-layer} for Fig.\ \ref{PR-curve_single-layer}, \texttt{python make\_plot2.py multilayer} for Fig.\ \ref{PR-curve_multi-layer}, \texttt{python \\make\_plot2.py transformer} for Fig.\ \ref{PR-curve_transformer}, and \texttt{python make\_plot2.py \\nested} for Fig.\ \ref{PR-curve_nested_validation}).

\section{Using the RNN Model to Predict \\ O-GlcNAcylation Sites} \label{Appendix_using_model}
Following the procedures set by Ref.\ \cite{Seber-and-Braatz-2023_2}, the Conda environment defining the specific packages and version numbers used in this work is available as \texttt{ANN\_environment.yaml} on \href{https://github.com/PedroSeber/O-GlcNAcylation_Prediction}{https://github.com/PedroSeber/\\O-GlcNAcylation\_Prediction}. To use our trained model, run the \\ \texttt{Predict.py} file as \texttt{python Predict.py <sequence> -t <threshold> -bs <batch\_size>}.

Alternatively, create an (N+1)x1 .csv with the first row as a header (such as "Sequences") and all other N rows as the actual amino acid sequences, then run the Predict.py file as \texttt{python Predict.py \\ <path/to/file.csv> -t <threshold> -bs <batch\_size>}. Results are saved as a new .csv file.

\section{List of Hyperparameters Used in This Work} \label{appendix_hyperparameters}
\begin{enumerate}
    \item For the transformer encoder models: embed sizes (ES) \(=\) \{24, 60, or 120\}, learning rates \(=\) \{1\(\times 10^{-3}\) or 1\(\times 10^{-4}\)\}, \{2, 4, 6, or 8\} stacked encoder cells, cell hidden sizes (widths) \(=\) \{ES/4, ES/2, ES, ES\(\times\)2, ES\(\times\)3, ES\(\times\)4, ES\(\times\)5\}, \{2, 3, 4, 5, or 6\} attention heads (when available in a given ES), post-encoder MLP sizes \(=\) \{0, ES/4, ES/2, ES\}, and window sizes of \{5, 10, 15, or 20\} were used. Training is done in 100 epochs with an AdamW optimizer with a weight decay parameter \(\lambda =\) \{0, 1\(\times 10^{-2}\), 5\(\times 10^{-2}\), or 1\(\times 10^{-1}\)\} and a cosine scheduler with a minimum learning rate equal to 1/10 of the original learning rate with 10 epochs of warm-up.
    \item For the ProtBERT finetuning: learning rates \(=\) \{1\(\times 10^{-3}\), 5\(\times 10^{-4}\), 1\(\times 10^{-4}\), 5\(\times 10^{-5}\)\} and window sizes of \{10, 15, 20, 25, 30\} were used. Training is done in 40 epochs with early stopping set with a patience of 2 epochs, and a cosine scheduler with a minimum learning rate of 0 with 5 epochs of warm-up.
    \item For the RNN (LSTM) models: RNN hidden sizes \(=\) \{225 to 1575 in multiples of 75\} for the first cell, \(=\) \{0 to 300 in multiples of 75\} for the second cell, and \(=\) \{0 to 225 in multiples of 75\} for the third and fourth cells, learning rates \(=\) \{1\(\times 10^{-3}\), 5\(\times 10^{-4}\), 1\(\times 10^{-4}\)\}, post-LSTM MLP sizes \(=\) \{37 and 75 to 825 in multiples of 75\}, and window sizes of \{20\} were used. Training is done in 70 epochs with an AdamW optimizer with a weight decay parameter \(\lambda =\) \{0, 1\(\times 10^{-2}\), or 2\(\times 10^{-2}\)\} and a cosine scheduler with a minimum learning rate equal to 1/10 of the original learning rate with 20 epochs of warm-up.
    \item For the weighted CE loss, the positive-class weights used were \{15, 20, 25, 30\}. For the focal loss, the \(\alpha\) values used were \{0.9375, 0.95, 0.99, 0.999, 0.9999, 0.99999, 0.999999\} and the \(\gamma\) values used were \{0, 1, 2, 3, 5, 10, 15, 20, 30, 45, 50\}. For the weighted focal differentiable MCC, the weights W used were \{1, 2, 3, 4, 5, 10, 15, 20\} and the \(\gamma\) values used were \{1, 2, 3, 3.5, 4, 4.5, 5\}.
\end{enumerate}
The transformer encoder and RNN architectures were initialized with the Kaiming uniform initialization \cite{He-etal-2015} with \(a = \sqrt{5}\). Cosine scheduling is used because it has achieved significant results with imbalanced datasets \cite{Kukleva-etal-2023,Mishra-etal-2019}, and it was also used by Ref.\ \cite{Seber-and-Braatz-2023_2}.

\section{Supplemental Results}
\setcounter{figure}{0}
\makeatletter 
\renewcommand{\thefigure}{S\@arabic\c@figure}
\makeatother
\setcounter{table}{0}
\makeatletter 
\renewcommand{\thetable}{S\@arabic\c@table}
\makeatother

\subsection{Transformer encoder models from this work surpass previous transformer models, but are inferior to RNNs} \label{Appendix_transformers}
\begin{figure}[H]
    \hspace{-12ex}
    \centerline{
        \includegraphics[width=1.85\textwidth]{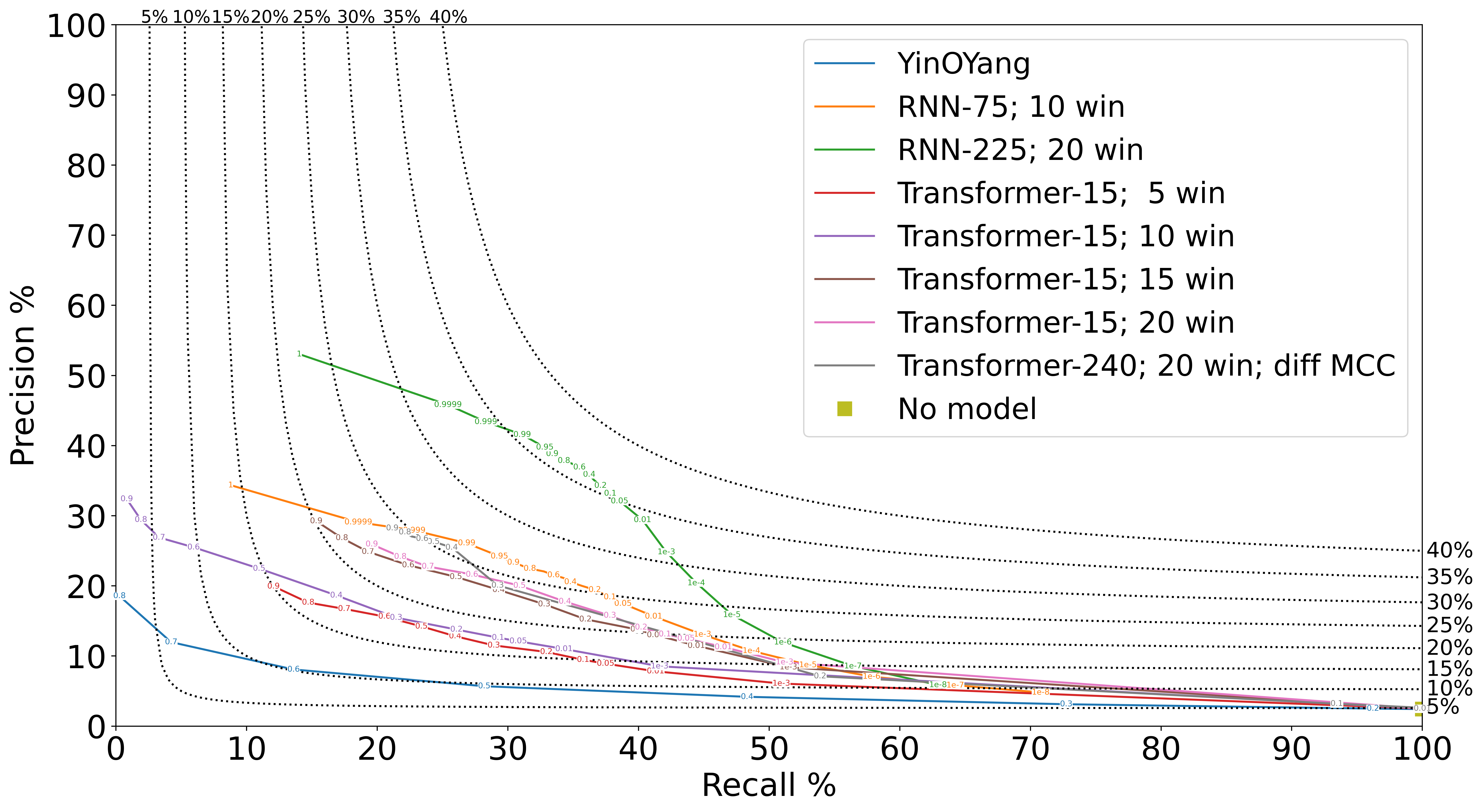} }
    \caption{P-R curves for O-GlcNAcylation site prediction models tested on the modified dataset of Ref.\ \cite{Wulff-Fuentes-etal-2021}. Labels, curve descriptions, and baseline metrics are as in the captions of Table \ref{Metrics_transformer} and Fig.\ \ref{PR-curve_finetune}.}
    \label{PR-curve_transformer}
\end{figure}

The primary goal of this subsection is to compare transformer models with the best RNN from Ref.\ \cite{Seber-and-Braatz-2023_2}, labeled in this work as ``RNN-225'' or ``RNN-225; CE'', using the modified dataset of Ref.\ \cite{Wulff-Fuentes-etal-2021} (Section \nameref{MnM-dataset}). While this work's transformer models surpass models published before Ref.\ \cite{Seber-and-Braatz-2023_2} and achieve an F\(_1\) score equal to 24.31\% and an MCC equal to 22.49\% (Table \ref{Metrics_transformer}), transformers exhibit inferior recall at the same precision level (Fig.\ \ref{PR-curve_transformer}) than the RNNs from Ref.\ \cite{Seber-and-Braatz-2023_2}, leading to inferior F\(_1\) and MCC metrics. It should be noted that the work of Ref.\ \cite{Pokharel-etal-2023} used an ensemble of three transformer models to predict O-GlcNAcylation and obtained considerably inferior F\(_1\) and MCC metrics than the RNN of Ref.\ \cite{Seber-and-Braatz-2023_2} and the transformer models in this work. 

A surprising result is how the performance of transformer encoder models does not change significantly with the hyperparameters used. Using large hidden sizes or adding another linear layer after the encoders slightly degrades the performance, while increasing the number of attention heads (to 4--5) and the number of stacked encoder layers (to 4) marginally improves the model. The most surprising result is the lack of significant improvement with increasing window size. As in Ref.\ \cite{Seber-and-Braatz-2023_2}, our models perform better with larger window sizes for sizes up to 20 (Fig.\ \ref{PR-curve_transformer} and Table \ref{Metrics_transformer}). The performance of a transformer encoder with a window size = 5 is only marginally inferior to the RNN from Ref.\ \cite{Seber-and-Braatz-2023_2} using the same window size. However, whereas the RNNs' performances increase quite significantly with increasing window size, the performance of the transformers barely improves. The only significant performance increase comes from increasing the window size from 10 to 15, but little performance is gained from further increasing this window size to 20 (Fig.\ \ref{PR-curve_transformer}).

We hypothesized that changing the loss function can lead to improvements in the models' metrics, as the weighted CE is optimizing a function that is only correlated with F\(_1\) scores and MCC. Four other loss functions (as described in Section \nameref{MnM-loss}) are also used to cross-validate transformer models through the same procedure used with the weighted CE function. The Focal, Lov{\'a}sz, and differentiable F\(_1\) losses fail to achieve meaningful results. In the best-case scenarios, models trained with these loss functions can achieve F\(_1\) scores equal to or slightly above 4.74\%, equivalent to treating all (or almost all) sequences as positive. The differentiable F\(_1\), in particular, frequently gets stuck at this local F\(_1\) maximum. These models have an MCC \(\approx 0\), as they are no better than random classifiers. However, a transformer model trained with the weighted focal differentiable MCC was able to surpass the performance of the transformers trained with the weighted CE loss (Table \ref{Metrics_transformer} and Fig.\ \ref{PR-curve_transformer}), improving the F\(_1\) score to 25.59\% and the MCC to 23.67\%. Nevertheless, these metrics remain inferior to the previous state-of-the-art RNN model.

\begin{table}[H]
    \caption{Performance metrics (in \%) for the highest F$_1$ score point of models tested on the modified dataset of Ref.\ \cite{Wulff-Fuentes-etal-2021}. Models ``Transformer-\#'' come from this work. The ``RNN-225'' model \cite{Seber-and-Braatz-2023_2} is the previous state-of-the-art model. The number after the RNN model is the LSTM module size used; the number after each transformer model is the hidden size used; WS is the window size on each side of the central S/T. Metrics for the model ``YinOYang'' \cite{Gupta-and-Brunak-2002} were obtained by Ref.\ \cite{Seber-and-Braatz-2023_2}. Baseline (threshold = 0) metrics are precision = 2.44\%, F\(_1\) score = 4.76\%, and MCC = 0\%.}
    \label{Metrics_transformer}
    \centerline{\begin{tabular}{| c | c c c c |}
    \hline
    \multirow{2}{*}{Metric} & \multicolumn{4}{c |}{Transformer-15; CE (This Work)} \\
    & WS = 5 & WS = 10 & WS = 15 & WS = 20 \\
    \hline
    Best Threshold   &  0.60 &  0.20 &  0.50 &  0.50 \\
    Recall (\%)      & 20.54 & 26.06 & 26.02 & 30.89 \\
    Precision (\%)   & 15.63 & 13.77 & 21.28 & 20.04 \\
    F$_1$ Score (\%) & 17.75 & 18.02 & 23.41 & 24.31 \\
    MCC (\%)         & 15.50 & 16.08 & 21.36 & 22.49 \\
    \hline
    \end{tabular} }
    \vspace{1ex}
    \centerline{\begin{tabular}{| c | c | c | c |}
    \hline
    \multirow{2}{*}{Metric} & \thead{Transformer-240; diff MCC \\ (This Work)} & \thead{RNN-225; CE \\ (Ref.\ \cite{Seber-and-Braatz-2023_2})} & \multirow{2}{*}{\thead{YinOYang \\ (Ref.\ \cite{Gupta-and-Brunak-2002})}}\\
    & WS = 20 & WS = 20 &  \\
    \hline
    Best Threshold   & 0.40  & 0.60  &  0.60 \\
    Recall (\%)      & 25.70 & 35.47 & 13.59 \\
    Precision (\%)   & 25.48 & 36.90 &  8.08 \\
    F$_1$ Score (\%) & 25.59 & 36.17 & 10.13 \\
    MCC (\%)         & 23.67 & 34.57 &  7.57 \\
    \hline
    \end{tabular} }
\end{table}

\subsection{Nested validation of the two-layer RNN models} \label{Appendix_nested_validation}

\begin{figure}[H]
    \hspace{-12ex}
    \centerline{
        \includegraphics[width=1.85\textwidth]{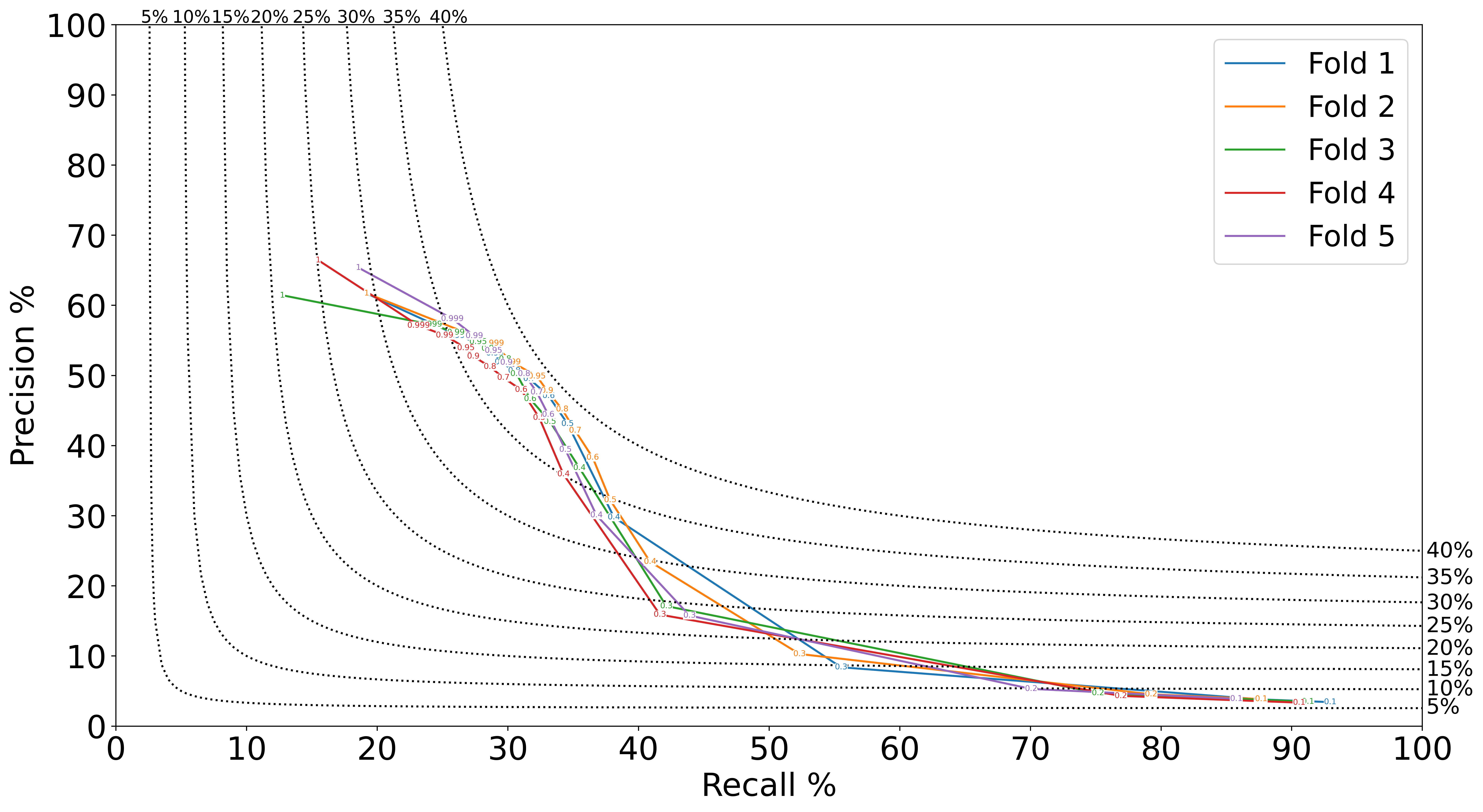} }
    \caption{Precision-recall curves for the ``RNN-[600,75]; diff MCC'' model after nested validation on the modified dataset of Ref.\ \cite{Wulff-Fuentes-etal-2021}. The black dotted lines are F$_1$ score isolines, as labeled on the top and right sides of the figure.}
    \label{PR-curve_nested_validation}
\end{figure}

\begin{table}[H]
    \caption{Recall, precision, F$_1$ score, and MCC metrics (in \%) for the point with highest F$_1$ score for the ``RNN-[600,75]; diff MCC'' model after nested validation on the modified dataset of \cite{Wulff-Fuentes-etal-2021}. Baseline metrics are as in the caption of Table \ref{Metrics_transformer}.}
    \label{Metrics_nested_validation_MCC}
    \centerline{
    \begin{tabular}{| c | c c c c c | c |}
    \hline
    Metric         & Fold 1 & Fold 2 & Fold 3 & Fold 4 & Fold 5 & Mean (\(\pm \sigma\)) \\
    \hline
    Best Threshold   &  0.60 &  0.95 &  0.70 &  0.60 &  0.80 & N/A \\
    Recall (\%)      & 33.11 & 32.22 & 30.66 & 31.01 & 30.66 & 31.53\(\pm\)0.98 \\
    Precision (\%)   & 47.10 & 49.88 & 50.24 & 47.96 & 50.24 & 49.08\(\pm\)1.30 \\
    F$_1$ Score (\%) & 38.88 & 39.15 & 38.08 & 37.66 & 38.51 & 38.46\(\pm\)0.54 \\
    MCC (\%)         & 38.20 & 38.93 & 38.07 & 37.38 & 38.44 & 38.20\(\pm\)0.51 \\
    \hline
    \end{tabular} }
\end{table}

\begin{table}[H]
    \caption{Recall, precision, F$_1$ score, and MCC metrics (in \%) for the point with highest F$_1$ score for the ``RNN-[450,75]; CE'' model after nested validation on the modified dataset of \cite{Wulff-Fuentes-etal-2021}. Baseline metrics are as in the caption of Table \ref{Metrics_transformer}.}
    \label{Metrics_nested_validation_CE}
    \centerline{
    \begin{tabular}{| c | c c c c c | c |}
    \hline
    Metric         & Fold 1 & Fold 2 & Fold 3 & Fold 4 & Fold 5 & Mean (\(\pm \sigma\)) \\
    \hline
    Best Threshold   &  0.90 &  0.90 & 0.999 &  0.95 &  0.95 & N/A \\
    Recall (\%)      & 36.69 & 39.24 & 36.48 & 36.10 & 35.87 & 36.88\(\pm\)1.22 \\
    Precision (\%)   & 41.02 & 38.98 & 40.69 & 38.30 & 39.31 & 39.66\(\pm\)1.03 \\
    F$_1$ Score (\%) & 38.73 & 39.11 & 38.47 & 37.17 & 37.51 & 38.20\(\pm\)0.74 \\
    MCC (\%)         & 37.31 & 37.60 & 37.06 & 35.70 & 36.05 & 36.74\(\pm\)0.74 \\
    \hline
    \end{tabular} }
\end{table}

\section{Computational Resources Used} \label{Appendix_computational_resources}
All experiments were done in a personal computer equipped with a 13th Gen Intel\textsuperscript{\textregistered} Core™ i5-13600K CPU, 64 GB of DDR4 RAM, and an NVIDIA GeForce RTX 4090 GPU.

\end{document}